\documentclass[letterpaper]{article}

\usepackage{natbib,alifeconf}  %% The order is important
\usepackage{url,hyperref,cleveref}
\usepackage{booktabs}
\usepackage{float}
\usepackage{censor}
\usepackage{csquotes}

\title{Coralai: Intrinsic Evolution of Embodied Neural Cellular Automata Ecosystems}

\author{
    Aidan Barbieux$^{1}$ \and
    Rodrigo Canaan$^{1}$
    \mbox{}\\
    $^1$California Polytechnic State University, San Luis Obispo \\
    aidanbx@email.com
} % email of corresponding author

\begin{document}

\maketitle

% Understanding the principles of resilient ecosystems is increasingly relevant to both the natural and digital worlds.
% Unfortunately, simulating ecosystems can be computationally intensive and narrow in scope, preventing the scaling that
% might be crucial to their properties. We introduce Coralai as
% a means of quickly developing scalable ecosystems of neural
% cellular automata (NCA). NCA in Coralai are embodied via
% physics written with easy-to-code Taichi kernels and develop
% via objective-free evolution using PyTorch HyperNEAT in
% persistent environments. We provide an exploratory experiment implementing physics inspired by slime mold behavior, showcasing the emergence of competition between sessile and mobile organisms, cycles of resource depletion and
% recovery, and apparent symbiosis between diverse organisms.
% We then finish by outlining future work to quantitatively discover simulation parameters by measures of multi-scale complexity and diversity.

\begin{abstract}
    % Abstract length should not exceed 250 words
    This paper presents Coralai, a framework for exploring diverse ecosystems of Neural Cellular Automata (NCA). Organisms in Coralai utilize modular, GPU-accelerated Taichi kernels to interact, enact environmental changes, and evolve through local survival, merging, and mutation operations implemented with HyperNEAT and PyTorch. We provide an exploratory experiment implementing physics inspired by slime mold behavior showcasing the emergence of competition between sessile and mobile organisms, cycles of resource depletion and recovery, and symbiosis between diverse organisms. We conclude by outlining future work to discover simulation parameters through measures of multi-scale complexity and diversity. Code for Coralai is available at \url{https://github.com/aidanbx/coralai}, video demos are available at \url{https://www.youtube.com/watch?v=NL8IZQY02-8}. 
\end{abstract}

\section{Introduction}
Most, if not all, of life exists within diverse networks of organisms that interact with each other, physics, and their environment to collectively develop resilience. Simulating this complexity manually is impractical and limited by our understanding of the system. Conversely, the emergence of complex behavior from simple rules, as demonstrated by many Cellular Automata or Neural Soups, can be difficult to interpret, control, and compute \cite{ plantec_flow-lenia_2023, gregor_self-organizing_2021}. Neural Cellular Automata (NCA) enable efficient training of low-level rules to create systems which self-organize to a defined state and have shown success in a wide variety of tasks \citep{mordvintsev_growing_2022, pande_hierarchical_2023, stovold_neural_2023}. By leaving trained NCA running, minimal evolutionary properties emerge \cite{sinapayen_self-replication_2023}. By explicitly evolving rules to survive and interact with `physics,' they display more diverse, lifelike behaviors \cite{randazzo_biomaker_2023}. This shows significant potential to model the emergence of complex ecosystems, but current work is limited to specific behavior around plant-like growth.

Coralai is designed for the exploration of a wide variety of organism, environment, and physics configurations in evolved NCA ecosystems. Organisms in Corlai produce abstract output which can then be arbitrarily transformed by physics to enact change in the environment, interact with other organisms, and evolve. Experiments can be defined at a high level and arbitrary physics can be implemented using GPU-accelerated Taichi kernels \citep{hu_taichi_2019}. Interactive simulations run with simple visualizations in real-time and could be scaled to larger systems. These features provide the basis of a powerful experimental platform for studying the emergence of complex, resilient ecosystems.

\begin{figure}
    \centering
    \includegraphics[width=3.1in]{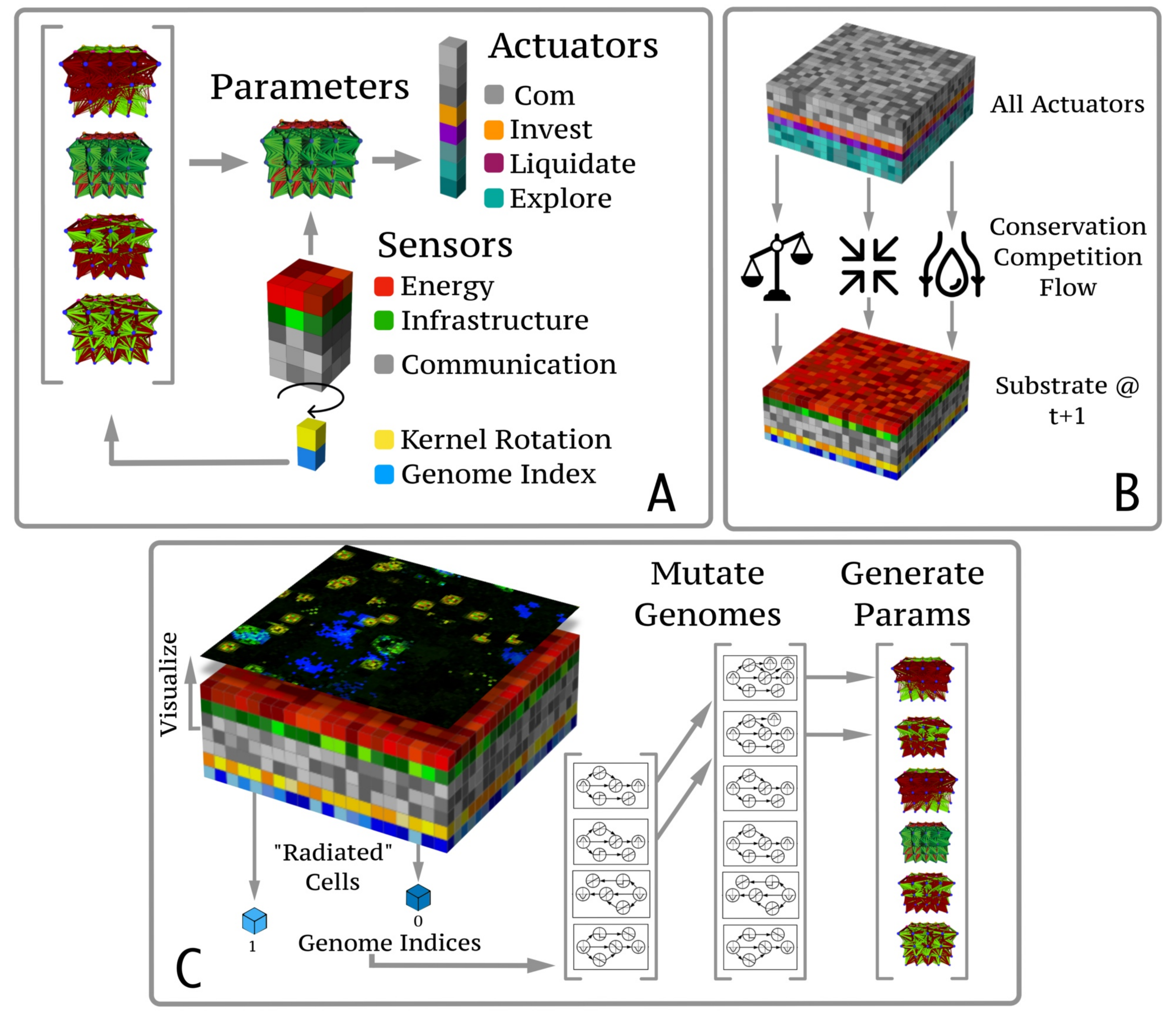}
    \caption{
        A single update in Coralai. (A): The sensing of substrate and production of actions. (B): Applying the actions of the entire system to the substrate via `physics'. (C): Visualization and radiation of the substrate resulting in mutation of CPPN genomes and generation of dense network parameters.
    }
    \label{methods_fig}
\end{figure}

\section{System Design}
% What needs to be said?
Each experiment in Coralai is defined by Substrate, Sensors, Actuators, Kernel, and Physics. The substrate is a tensor with width, height, and named channels/sub-channels. One such channel is $GenomeIndex$, which locates the neural network parameters governing the behavior of the cell. In discrete time steps, each cell is updated by sending a neighborhood (the Kernel) of channels (the Sensors) to the cell’s network to receive values for the Actuator channels, which are then transformed via Physics to produce changes in the substrate.

The networks of each cell are generated via HyperNEAT (Hypercube Based Neuroevolution of Augmented Topologies) \citep{stanley_hypercube-based_2009} implemented in PyTorch which uses small Compositional Pattern Producing Networks (CPPN) to generate the weights and biases of a larger network. Coralai uses only the mutation and crossover operators of NEAT, which are applied via the Physics.

\section{Experiment}
To test Coralai, we conducted exploratory experiments resembling the physical affordances exploited by slime molds. This was chosen for the well-studied behaviors slime molds can exhibit with simple physiologies, such as network optimization, transfer learning, and decision making \cite{ray_information_2019, vogel_direct_2016, vallverdu_slime_2017}, developing the work of \cite{barbieux_eincasm_2023}.

The main channels of the experiment are Infrastructure, Energy, and Communication (shown with 3 sub-channels in figure \ref{methods_fig}). Energy is added and removed from the substrate in a day/night cycle. Organisms can transform Energy into Infrastructure via Invest and Liquidate actuators. Communication channels are simply normalized and activated via a sigmoid function. The kernel is a Moore's neighborhood with offsets ordered by increasing angular distance to a Rotation channel of each cell. 

Organisms spread via Explore actuators matched to each offset in the kernel. Exploration spreads infrastructure to a neighboring cell determined by the maximal actuator. If the quantity of the exploratory infrastructure exceeds that of the explored cell, the explored cell adopts the genome of the original cell and the rotation of the exploration. Otherwise, the infrastructure is taken while the current genome and rotation are maintained. If multiple cells explore a given cell, the one with maximal exploratory investment wins.

\begin{figure}
    \centering
    \includegraphics[width=3.1in]{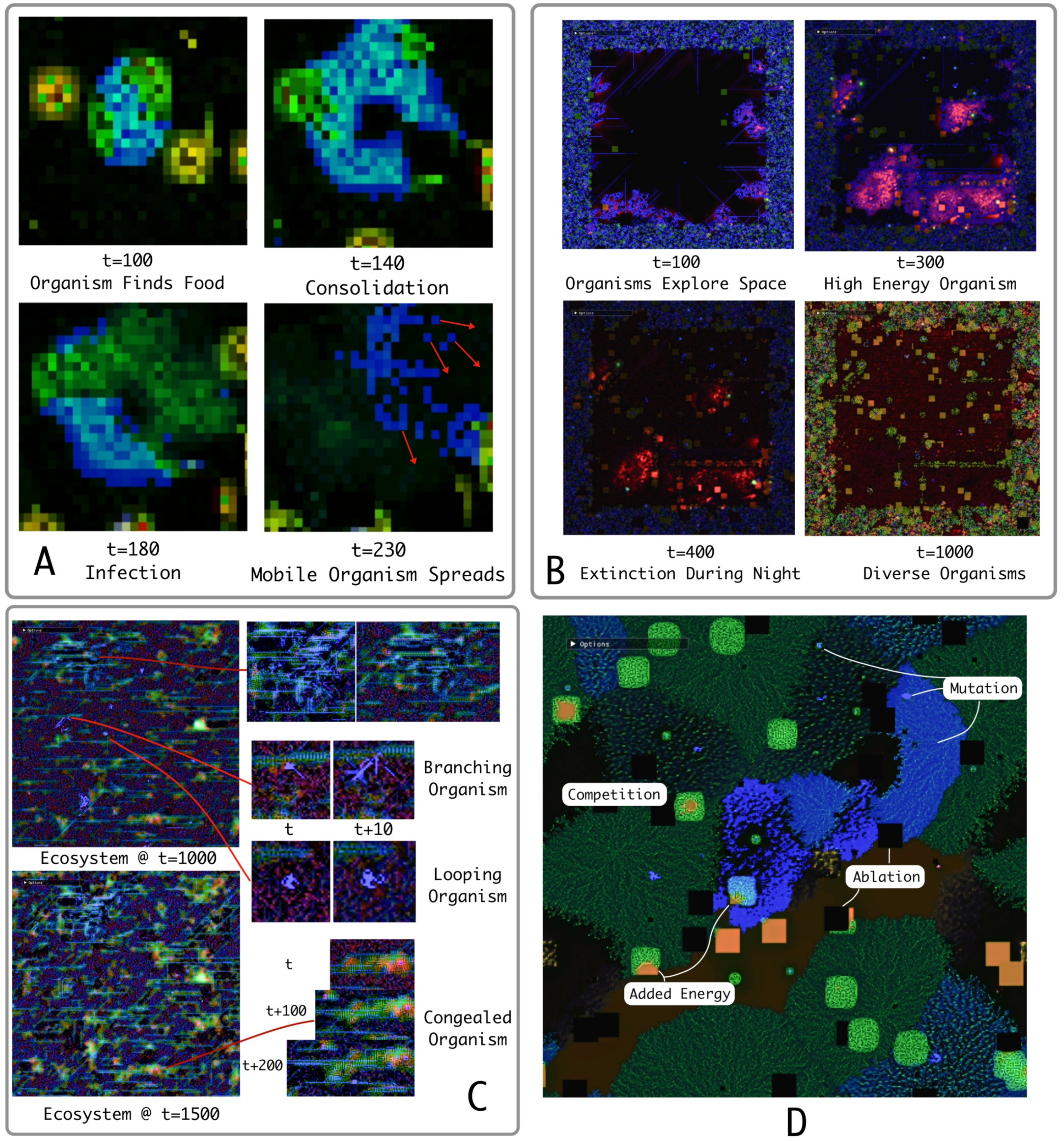}
    \caption{
        Notable experiment runs. Visual observation shows competition between mobile and sessile organisms in (A), extinction and replacement in (B), a variety of organisms stabilized in an ecosystem in (C), and competing slime-mold-like organisms in (D). Energy is displayed as red, infrastructure by green, and genome/species by blue. 
    }
    \label{results_fig}
\end{figure}

We ran exploratory simulations modulating substrate size, initial conditions, and hyper-parameters. We observed a wide variety of behaviors and patterns requiring further quantitative analysis. Figure \ref{results_fig} provides noteworthy examples showing mobile and sessile organisms, dominance overcome by diversity, and an ecosystem at equilibrium. We additionally observed cycles of greedy organisms consuming excess energy and causing mass-extinction followed by diversity and resilience and networked organisms spreading via spore-like cells upon depletion of an energy pocket.

Simulations of 1024x1024 cells with 512 unique organisms ran at 20 FPS on an M1 chip. Runs were limited to manual set up and observation with various initial conditions and physical parameters.

    \section{Discussion and Future Work}
    Evolving diverse Neural Cellular Automata to collectively survive physics provides an easy-to-use and powerful basis to study artificial life. However, due to the large parameter spaces and possible physics in Coralai, intelligent search strategies should be employed. To move beyond manual observation and implementation of physics, we suggest meta-evolution of both parameters and physical rules to optimize known indicators of life.
    
    One promising metric is multi-scale complexity (MSC) to detect self-organizing criticality, a crucial component of living systems \citep{bagrov_multiscale_2020, bak_self-organized_1989}. MSC calculation via renormalization-groups (but not meta-evolution) is implemented in Coralai and described in more detail by \cite{barbieux_coralai_2024}. We also suggest ecological measures of diversity \citep{eysenbach_diversity_2018}. Coralai can use NEAT's measure of genetic similarity \citep{stanley_evolving_2002} and tracks phylogeny to enable this.
    
    Finally, we suggest future experiments that cater to niche formation, promoted through physics for obstacles, local weather patterns, `cross-pollination’ between simulations, among other techniques.

\footnotesize
\bibliographystyle{apalike}
\bibliography{main} % replace by the name of your .bib file

\end{document}